\documentclass[letterpaper]{article} 
\usepackage[]{aaai25}  
\usepackage{times}  
\usepackage{helvet}  
\usepackage{courier}  
\usepackage[hyphens]{url}  
\usepackage{graphicx} 
\usepackage{natbib}  
\usepackage{caption} 
\usepackage{algorithm}
\usepackage{algorithmic}
\usepackage{amsmath}
\usepackage{mathtools}

\usepackage{amsmath,amsfonts,bm}

\def\eqref#1{equation~\ref{#1}}

\def\1{\bm{1}}

\def\rvm{{\mathbf{m}}}

\def\rvq{{\mathbf{q}}}

\DeclareMathAlphabet{\mathsfit}{\encodingdefault}{\sfdefault}{m}{sl}
\SetMathAlphabet{\mathsfit}{bold}{\encodingdefault}{\sfdefault}{bx}{n}

\def\gL{{\mathcal{L}}}

\def\gO{{\mathcal{O}}}

\usepackage{newfloat}
\usepackage{listings}
\DeclareCaptionStyle{ruled}{labelfont=normalfont,labelsep=colon,strut=off} 
\floatstyle{ruled}
\newfloat{listing}{tb}{lst}{}
\floatname{listing}{Listing}
\title{\textsc{CMT}: A Memory Compression Method for Continual Knowledge Learning of Large Language Models}
\author{
    Dongfang Li,
    Zetian Sun,
    Xinshuo Hu,
    Baotian Hu,
    Min Zhang
}
\affiliations{

    Harbin Institute of Technology (Shenzhen)\\
    crazyofapple@gmail.com

}

\usepackage{bibentry}
\usepackage{booktabs}
\usepackage{multirow}
\usepackage{microtype}
\usepackage{graphicx}
\usepackage{subfigure}
\usepackage{booktabs} 
\usepackage{duckuments}
\usepackage{multirow}
\usepackage{makecell}
\usepackage{enumitem}
\usepackage{xcolor} 

\usepackage[switch]{lineno}

\definecolor{orange}{RGB}{255,165,0}
\usepackage{colortbl}
\definecolor{darkblue}{rgb}{0,0.08,0.45}
\definecolor{Gray}{gray}{0.9}

\newcommand{\answerTODO}[1][]{\textcolor{red}{\bf [TODO]}}
\newcommand{\justificationTODO}[1][]{\textcolor{red}{\bf [TODO]}}

\begin{document}
\maketitle
\begin{abstract}
Large Language Models (LLMs) need to adapt to the continuous changes in data, tasks, and user preferences. Due to their massive size and the high costs associated with training, LLMs are not suitable for frequent retraining. However, updates are necessary to keep them in sync with rapidly evolving human knowledge. To address these challenges, this paper proposes the \textbf{C}ompression \textbf{M}emory \textbf{T}raining (\textbf{CMT}) method, an efficient and effective online adaptation framework for LLMs that features robust knowledge retention capabilities.
Inspired by human memory mechanisms, CMT compresses and extracts information from new documents to be stored in a memory bank. When answering to queries related to these new documents, the model aggregates these document memories from the memory bank to better answer user questions. The parameters of the LLM itself do not change during training and inference, reducing the risk of catastrophic forgetting. To enhance the encoding, retrieval, and aggregation of memory, we further propose three new general and flexible techniques, including memory-aware objective, self-matching and top-$k$ aggregation. Extensive experiments conducted on three continual learning datasets (i.e., StreamingQA, SQuAD and ArchivalQA) demonstrate that the proposed method improves model adaptability and robustness across multiple base LLMs (e.g., +4.07 EM \& +4.19 F1 in StreamingQA with \texttt{Llama-2-7b}).
\end{abstract}

\section{Introduction}
\label{sec:intro}
Large language models (LLMs) have become the core of natural language processing (NLP)~\cite{DBLP:journals/corr/abs-2307-09288,DBLP:journals/corr/abs-2303-08774,DBLP:journals/corr/abs-2312-11805}. The current challenge is how these LLMs adapt to rapidly changing world knowledge, especially in the context of increasing new data and growing model complexity~\cite{DBLP:journals/corr/abs-2404-16789}. Typically, LLMs are trained on static and pre-defined datasets. For example, the Llama-3.1 model is an open-source large language model by Meta, with a training dataset over 15 trillion tokens~\cite{2023-Llama2}. However, in practical applications, language usage habits, information content, and user needs are all dynamically changing~\cite{DBLP:journals/corr/abs-2402-01364}. On the other hand, once training is complete, the model becomes fixed, and the cost and  computational demands of retraining or incremental pre-training is extremely high. For example, the GPT-3 model has 174.6 billion parameters, and retraining it once requires approximately 3640 PF-days of computing power (i.e., performing 10 quadrillion calculations per second for 3640 days)~\cite{DBLP:conf/nips/BrownMRSKDNSSAA20}. Therefore, how to adapt downstream  tasks \textit{effectively} and \textit{efficiently} with updating the model with the new knowledge while retaining the existing knowledge has become an important and urgent topic.
\begin{figure}[t]
\centering
\includegraphics[width=0.4\textwidth]{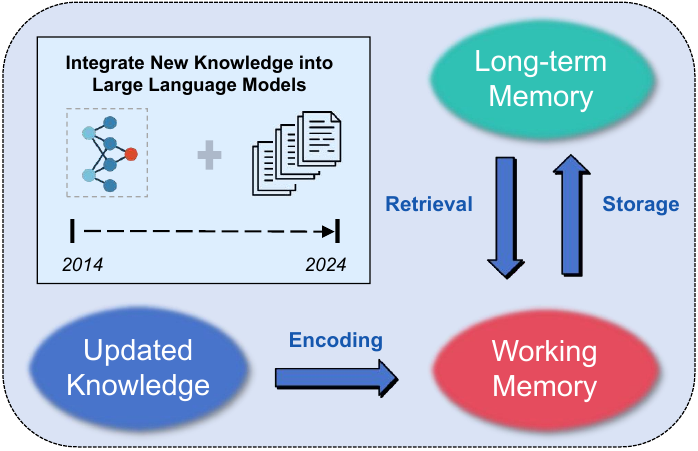} 
\caption{In general, memory can be divided into three stages: (1) \textit{encoding} involves reorganizing and transforming external information; (2) \textit{storage} entails hierarchically categorizing and preserving information in long-term memory; (3) \textit{retrieval} extracts information from long-term memory. We draw on this process to address the continual learning of LLMs when dealing with online streaming documents.}
\label{fig:motivation}
\vspace{-4mm}
\end{figure}

To address these challenges, existing continual learning methods dynamically update new incremental knowledge through techniques such as data replay and incremental task parameters while balancing the generalization of new and old knowledge~\cite{schwarz2018progress,riemer2018learning,shi2024unified}. However, data replay and incrementally adding task parameters bring non-negligible computational overhead. Additionally, methods like model editing only provide patches to the model, resulting in poor generalization ability and even causing collapse~\cite{DBLP:conf/emnlp/YaoWT0LDC023}. While it is possible to perform thousands of edits simultaneously, scalability is low when updating a large amount of knowledge in large models, and catastrophic forgetting can occur. 
On the other hand, memory is the foundation of human intelligence while humans use memory to achieve continual learning. As shown in Figure~\ref{fig:motivation}, memory impact on intellectual activities such as learning, abstraction, association, and reasoning in the human brain spans three stages: (1) Encoding: It involves reorganizing and transforming external information. The efficiency of learning depends on the strategies used for memory encoding~\cite{pyc2010testing}. Strategies such as multi-channel encoding and contextual association can significantly enhance learning outcomes. (2) Storage: Information is stored hierarchically and categorized in long-term memory. Retaining learned content makes subsequent learning more efficient~\cite{bjork1994memory}.
(3) Retrieval: It involves extracting and aggregating information from long-term memory. It consolidates memory storage, stimulates meta-cognitive abilities, and promotes reasoning, abstraction, and association~\cite{karpicke2008critical}. Hence, how to draw on human memory mechanisms to the continual learning process has become an interesting and possible direction for adapting to changing world knowledge of LLMs.

To this end, we introduce the \textbf{C}ompression \textbf{M}emory \textbf{T}raining (\textbf{CMT}) method that encodes knowledge extracted from new documents into a dynamic memory bank within its latent space, serving as long-term memory for subsequent retrieval and aggregation.
The core idea is to freeze the parameters of the LLM itself and construct a memory-based module that learns to automatically encode and collect relevant information. Specifically, we first utilize an instantiable compressor to compress information from new documents into compact representations, which are cached to maximize the performance of the LLM on unseen tasks. Different from~\citet{mac}, this representation is generated through memory tokens with decoder-only model representing compressed knowledge, resulting in a memory bank that is less redundant than traditional knowledge bases in retrieval-based methods or contexts in prompting compression methods. Thus, during online adaptation, each document stream instance is stored in the memory bank. It allows contexts to be pre-computed offline once and reducing the LLM’s computational costs at inference.
Next, we learn to aggregate representations (i.e., memory) in the feature space into a single representation based on the given query, which is then mapped into cached key-value pairs within each transformer layer of the LLM. 
To ensure the effectiveness and scalability of CMT, we further propose three training and inference techniques corresponding to the encoding, retrieval and aggregation stages of memory respectively: (1) memory-aware objective; (2) self-matching; and (3) top-$k$ aggregation.
The evaluation of CMT focuses on several key aspects: (1) Integration of new knowledge. The model’s performance is assessed with downstream QA tasks, where CMT demonstrates substantial improvements over existing methods, indicating the superiority of supplementing LLMs with CMT; (2) Knowledge retention. CMT is evaluated on knowledge retention experiments under scenarios with different numbers of adapted documents, showcasing its ability to recall knowledge; (3) Robustness. We use the proportion of unrelated documents as a measure to test the model’s performance in the presence of irrelevant interference. The results show that CMT outperforms competitive baselines, demonstrating superior robustness. 

Our contributions are summarized as follows: 
\begin{itemize}
    \item We introduce CMT that incorporates an integrated memory bank within the latent space to address the challenges of continual learning of LLMs. 
    \item  To utilize encoded memory more efficiently, we further propose three effective training and inference strategies.
    \item  CMT demonstrates competitive performance across three benchmarks and knowledge retention settings, showcasing its versatility, effectiveness, and robustness.
\end{itemize}

\begin{figure*}[t]
\centering
\includegraphics[width=1\textwidth]{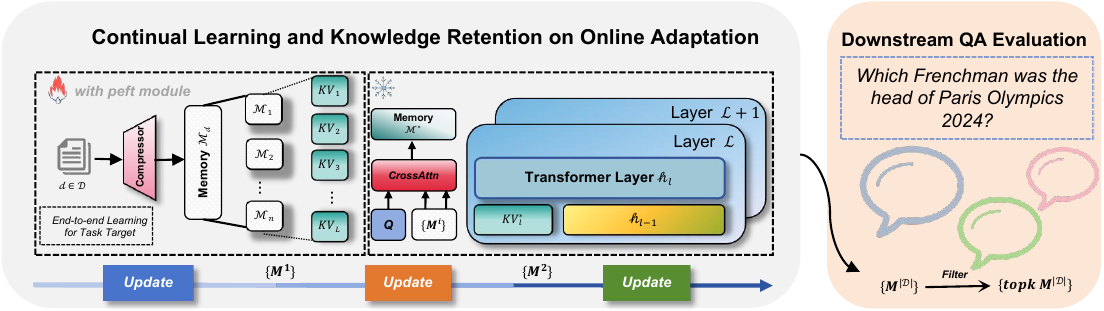} 
\caption{Illustration of compression memory training method. During the adaptation process, each document to be learned will be compressed into the dense vector by a compressor, and these vectors $\{M^{|\textit{D}|}\}$ will be aggregated through the cross attention mechanism and sent to LLMs together with the question for answer output. The training goal is the accuracy of the downstream task answer. In the online adaptation stage, all documents will be compressed into vectors and then filtered and aggregated.}
\vspace{-4mm}
\label{fig1}
\end{figure*}
\section{Related Work}
\label{sec:related}

\paragraph{Memory-Augmented Models} Memory-augmented models are not a new concept. Early memory networks introduced computational methods to store contextual information in limited space, thereby enhancing inference efficiency~\cite{weston2015memory,E2EMN,ba2016using}. Following this, Memory Transformer~\cite{memoryTransformer} and RMT~\cite{RMT} proposed adding memory tokens when reading contexts. However, expanding memory and incorporating information without disrupting the model's original capabilities remains a long-term challenge~\cite{kNNLM,DBLP:conf/emnlp/ZhongLC22,ret-llm,Memory-Enhanced-Transformer,MemoryBank,LongMEM,yang2024text}. Recent research has also focused on compressing prompts to enhance LLM inference efficiency~\cite{wingate2022prompt,snell2022learning,phang2023hypertuning}. For instance, AutoCompressor~\cite{chevalier2023adapting} and ICAE~\cite{gecontext} propose auto-encoding methods for compressing contexts into soft embeddings. Gisting~\cite{mu2024learning} introduces learnable tokens to compress context information within attention hidden states. Moreover, several improvements to transformers have demonstrated the benefits of equipping LLMs with external, controllable memory (e.g., MemoryLLM)~\cite{DBLP:journals/corr/abs-2312-03414, hmt, memoryllm}. However, these methods have not yet been applied to continual knowledge learning for existing LLMs as they typically require training from scratch and rely on inflexible and non-reusable implementations.

\paragraph{Continual Learning of LLMs} Continual learning aims to integrate LLMs into dynamic data distributions, task structures, and user preferences without significantly degrading performance in learned domains~\cite{DBLP:journals/corr/abs-2406-06391,DBLP:journals/corr/abs-2404-16789}. This involves sequentially training models on a series of tasks with the goal of maintaining performance across all tasks~\cite{kirkpatrick2017overcoming,li2017learning,riemer2018learning,buzzega2020dark}. During training, models often have limited or no access to previous data, making it challenging to retain past knowledge since optimization constraints from  previous data are absent during current-task learning~\cite{li2017learning,buzzega2020dark,smith2023closer,shi2024unified}. This challenge, known as \emph{catastrophic forgetting}~\cite{mccloskey1989catastrophic}, has been a central focus  since its inception.
Over the years, researchers have explored various techniques to mitigate forgetting in models. These include replay-based methods~\cite{schwarz2018progress,riemer2018learning,shi2024unified}, parameter regularization~\cite{kirkpatrick2017overcoming,ritter2018online,aljundi2018memory,sprechmann2018memory}, and model architecture expansion~\cite{wang2022coscl}. Recently, in the context of continual learning for LLMs, the challenge has shifted from storage efficiency to computational efficiency~\cite{DBLP:journals/corr/abs-2309-14763,DBLP:journals/pami/WangZSZ24}. In this paper, we focus on integrating memory systems into continual learning, enabling encoding, retrieval, and aggregation of knowledge without the need for expensive retraining.


\section{Method}
\label{sec:method}
\subsection{Task Formulation}
\label{Sec:taskformulation}
We consider the scenario where an outdated model, \( f_{\theta} \), is updated using an online stream of recent documents, \( D_{\text{test}} = \{x_i\} \). This process produces an updated model, \( f_{\theta+\triangle \theta} \), which is then evaluated against a set of queries, \( Q_{\text{test}} = \{q_i\} \), with corresponding labels, \( Y_{\text{test}} = \{y_i\} \). Each query \( q_i \) and its label \( y_i \) are derived from a distribution related to the corresponding document \( x_i \): \( q_i, y_i \sim p(q_i, y_i \mid x_i) \). For instance, \( q_i \) could be a question about some information in document \( x_i \), with \( y_i \) being the answer provided by the document.
A key constraint is that during the update process using \( D_{\text{test}} \), we do not have access to \( Q_{\text{test}} \). Thus, our methodology for updating \( f_{\theta} \) must be general rather than query-specific. 
To make this problem tractable, we assume the availability of an additional corpus of documents \( D_{\text{train}} \) and corresponding query samples \( Q_{\text{train}} \) and labels \( Y_{\text{train}} \) generated by a similar process to \( Q_{\text{test}}, Y_{\text{test}} \). This training set enables learning the types of queries that may be of interest, guiding us on how to update our model to optimize performance on test queries while minimizing disruption to its prior knowledge and behaviors. We define the adaptation process involving \( D_{\text{train}} \), \( Q_{\text{train}} \), and \( Y_{\text{train}} \) as the \textit{learning} phase, while the process involving \( D_{\text{test}} \) is referred to as the \textit{online adaptation} phase.
We next describe the method that adjusts the learning phase to more efficiently update our base model on the test stream of documents \( D_{\text{test}} \).

\subsection{CMT: Compression Memory Training}

Our goal is to efficiently adapt given LLMs to unseen knowledge while retaining previously learned knowledge, whether from the original pre-training stage or updates from documents in a stream of new data. To achieve this, we designed a learning method called \textbf{C}ompression \textbf{M}emory \textbf{T}raining shown in Figure~\ref{fig1}. During the online adaptation phase, it only requires a single forward pass to compress the documents knowledge into memory, which avoids the cost of gradient computation. Here, we first introduce the general process of the method.

First, each document $d$ in the document set $D$ is compressed into condensed vectors $\boldsymbol{M}$ through the compressor $\Theta$. Then, these dense vectors corresponding to each document are further aggregated. The aggregated vectors are further mapped and input into the LLM $\Phi$ to be adapted in the form of cached key-value pairs. The LLM $\Phi$ to be adapted freeze their parameters during both the training and online adaptation phases, reducing the risk of catastrophic forgetting during continuous updating. The parameters to be learned include memory encoding, storage, and mapping parts. The entire network is trained during the learning phase, with the learning objective being to better answer \( Q_{\text{train}} \).

\paragraph{Compression Memory} We introduce the method of compressing documents $D$ into condensed vectors. It aims to transform lengthy documents into concise, compact representations while striving to maintain the core semantics and integrity of the original knowledge.
We define a document $d \in D$ as $w=(w_1, w_2, \ldots, w_n, c_1, c_2, \ldots, c_k)$, where $w_i$ means the $i$-th token of document $d$, $c_j$ means the $j$-th soft virtual token adhere to this document, and $n$ is the number of actual tokens in document $d$. Let $\boldsymbol{e}(\cdot)$ represent the word embedding lookup in the LLM and $\boldsymbol{m}(\cdot)$ represent the learnable embeddings of soft tokens $c_1, c_2, \ldots, c_k$. A document compressor model $\Theta$ utilizes the document embeddings $\boldsymbol{e}(w) = (\boldsymbol{e}(w_1), \boldsymbol{e}(w_2), \ldots, \boldsymbol{e}(w_n))$ and the soft token embeddings $\boldsymbol{e}_{soft}(c) = (\boldsymbol{e}_{soft}(c_1), \boldsymbol{e}_{soft}(c_2), \ldots, \boldsymbol{e}_{soft}(c_k))$ to produce compact representations $\boldsymbol{M} = (\boldsymbol{m}_1, \boldsymbol{m}_2, \ldots, \boldsymbol{m}_k) \in \mathbb{R}^{k \times d}$ of the document $d$, where $k$ is the length of the compressed document and $k \ll n$. 
The condensed vectors $\boldsymbol{M}$ can replace the original context and be combined with other prompt embeddings $\boldsymbol{e}(p) = (\boldsymbol{e}(p_1), \ldots, \boldsymbol{e}(p_l))$ for input to an LLM $\Phi$.
The output $y = (y_1, \ldots, y_m)$ remains faithful to the content of the original context $w$. As illustrated in Figure \ref{fig1}, inspired by~\citet{gecontext}, the compressor can be instantiated as a series of cross-attention layers, pre-trained decoder models, and encoder-decoder models with a set of learnable soft tokens, termed condensed tokens. Here, the compressor utilizes document tokens and condensed tokens as inputs, leveraging a causal Transformer decoder to compress the document information into condensed vectors. We leave the application of these vector across different LLMs for future work.

\paragraph{Memory Aggregation} Given the memory bank of compressed documents $D$ represented as  $\{\boldsymbol{M}^i\}_{i=1}^{|D|}$, we aim to learn how to select most relevant information in the form of a transformation $\boldsymbol{M}^* \in \mathbb{R}^{k \times d}$ for a given input $q_i$. 
There are two feasible methods: (1) Retrieve one or multiple memory units and map them into the LLM space. For example, xRAG~\cite{xrag} addresses context aggregation from a multi-modal fusion perspective. It introduces a modality projector trained to directly project retrieved dense vectors into the LLM representation space. However, this approach has the risk of selecting incorrect memory units and requires a pre-training phase to learn how to resolve relationships among different memories. (2) Linearly interpolate multiple memory units, aggregate them by weights into a single memory unit, and map it into the LLM space. For example,~\citet{E2EMN} computes the weighted sum of the memory bank as the representative vector of the memory. The advantage of this method is that it can leverage ideas like attention mechanisms, model soups~\cite{wortsman2022model} and the mixture-of-experts method~\cite{moe} to filter and aggregate different memory units, maintaining permutation invariance of the memory units. However, such methods do not consider the relative position information of soft tokens within memory units during aggregation. Moreover, as we discussed in the task definition, the source of accurate information for answering question $Q_i$ is mostly related to the $i$-th document. Therefore, similar to~\citet{mac}, we select $\boldsymbol{M}^*$ using cross-attention blocks \citep{vaswani2017attention,kim2019attentive,xu2020metafun}, with the set aggregation network $\psi$:
\begin{equation}
\boldsymbol{M}^* = \psi \big(\Theta(q_i), \{\boldsymbol{M}^i\}_{i=1}^{|D|}\big)
\end{equation}
Here, for reasons of efficiency and consistency in the representation space, we use a document compressor $\Theta$ to compress the input $q_i$ (e.g., user query). Using an additional question encoder is left for future work.
As the vanilla cross-attention mechanism suffers from capturing the relative positional relationships among soft tokens within the document $d$. It implies that swapping any two tokens in the memory results in an identical condensed vector. Hence, in the aggregation process, we apply RoPE~\cite{ROPE} to represent the relative positional relations within the soft tokens. We only perform position embedding operations on query and key. And we allocate positional embeddings as if placing the soft tokens subsequent to the context tokens.
The RoPE embeddings $\mathrm{R}_i$ and $\mathrm{R}_j$ manifests the relative positional relationships through the inner product between $Q_{\mathrm{pos}}=\{\boldsymbol{q}_i\}:=\Theta(q_i)$ and $K_{\mathrm{pos}}=\{\boldsymbol{k}_{i}\}:=\{\boldsymbol{M}^k\}_{k=1}^{K}$:
\begin{equation}
    (\mathrm{R}_i\boldsymbol{q})^{T}(\mathrm{R}_j\boldsymbol{k})=\boldsymbol{q}^{T}\mathrm{R}_i^{T}\mathrm{R}_j\boldsymbol{k}=\boldsymbol{q}^{T}\mathrm{R}_{j-i}\boldsymbol{k}
\end{equation}
In this way, each soft token can recognize the relative positions relations of both intra- and inter-document soft tokens.

\paragraph{Alignment for LLM}
After obtaining $\boldsymbol{M}^* \in \mathbb{R}^{k \times d}$, we do not intend to use the memory as an embedding layer input to the LLM $\Phi$, as this does not fully leverage the memory to promote the association. Therefore, we design a network $\pi$ to map the original memory representation into cached key-value pairs (with the number being the actual tokens count). The purpose of $\pi$ is to perform self-attention on the memory tokens and use multiple multi-layer perceptrons to transform the memory tokens' features into actual tokens' features. Specifically, the module handles actual tokens by repeating memory tokens and recombines the processed features into the final output. Here, the actual tokens are defined as the new virtual tokens in each layer multiplied by the number of layers, then multiplied by 2 (i.e., key and value).

\paragraph{Training Objective} To train the memory embeddings $\boldsymbol{e}_{soft}$, the compressor $\Theta$, the aggregation networks $\psi$ and alignment module $\pi$, we optimize both networks end-to-end using the loss function $\gL$, which is the negative log-likelihood of the given label $y$:
\begin{equation}
    \gL = \min_{D_{\text{train}}, Q_{\text{train}},Y_{\text{train}}} 
    \frac{1}{N}\sum_{i=1}^{N}
    \mathcal{L}\big(\text{LM}_{\theta}(q_i; \pi(\boldsymbol{M}^{*})), y_i \big)
    \label{eq:train_obj}
\end{equation}
where $N$ is the number of the training queries and labels in $Q_{\text{train}}$. Note that we do not update the static LLM $\theta$ to avoid the risk of catastrophic forgetting by overwriting important parameters. It is important to train the model using the cross-entropy loss of the final QA task. We experimented with document auto-encoding pretraining tasks~\cite{gecontext}, with dividing it into two stages or using multi-task learning, but neither approach resulted in significant improvements.

\paragraph{Online Adaptation} 
After training the entire network on a given training corpus \( D_{\text{train}} \), \( Q_{\text{train}} \), and \( Y_{\text{train}} \), we introduce an online adaptation phase. Formally, the CMT processes a stream of test documents $D_{\text{test}}$ that are sequentially fed to the LLM. Considering that the task query $q^i_{\text{test}} \in Q_{\text{test}}$ is unavailable during the adaptation process, we first store the condensed representation of the document in the memory $\boldsymbol{M}=\{\Theta(d^{i}_{\text{test}})\}_{i=0}^{|D_{\text{test}}|}$  and later use the aggregation network to predict the modulation to adapt the LLM:
\begin{equation}
\hat{y}^i_{\text{test}}=\text{LLM}_{\theta}(q^i_{\text{test}}; \pi(\boldsymbol{M}^*))
\end{equation}
where $\boldsymbol{M}^* =\psi\big(\Theta(q^i_{\text{test}}), \boldsymbol{M}\big)$.

\subsection{Effective Learning Strategy}
\paragraph{Memory-Aware Conditional Objective}

To enhance the model's utilization of memory, we propose a training objective of contrastive ensemble between the logits. It enables the model to account for external knowledge which may not be aligned with the model's training data. Specifically, we adopt the vanilla logits from LLM as $l_{\theta}$ for the prior knowledge $l_{\theta}(y_i \mid \boldsymbol{q}_i)$. We demote this knowledge from the model's original output distribution via $\frac{l_{\theta}(y_i \mid \boldsymbol{M}^*, \boldsymbol{q}_i)}{l_{\theta}(y_i \mid \boldsymbol{q}_i)}$. We choose this formulation because it also represents the pointwise mutual information between the external knowledge from the document set $\boldsymbol{M}^*$ conditioned on $\boldsymbol{q}_i$. Optionally, one can adjust the original distribution by $\frac{l_{\theta}(y_i \mid \boldsymbol{M}^*, \boldsymbol{q}_i)}{l_{\theta}(y_i \mid \boldsymbol{M}^{-}, \boldsymbol{q}_i)}$, where $\boldsymbol{M}^{-}$ represents an explicit knowledge one wants to demote from (e.g., a set of other unrelated documents).
We interpolate this demotion $\frac{l_{\theta}(y_i \mid \boldsymbol{M}^*, \boldsymbol{q}_i)}{l_{\theta}(y_i \mid \boldsymbol{q}_i)}$ and the original output logits $l_\theta(y_i \mid \boldsymbol{M}^*, \boldsymbol{q})$ via a product-of-experts weighted by $\alpha$. We sample $y_i$ from the reweighted distribution:
\begin{equation}
    y_i \propto l_\theta(y_i \mid \boldsymbol{M}^*, \boldsymbol{q}_i) \left( \frac{l_{\theta}(y_i \mid \boldsymbol{M}^*, \boldsymbol{q})}{l_{\theta}(y_i \mid \boldsymbol{q}_i)} \right)^\alpha
\end{equation}
Further, we normalize it across all possible $y_i$:
\begin{equation}
    y_i \sim (1+\alpha) l_\theta(y_i \mid \boldsymbol{M}^*, \boldsymbol{q}_i) - \alpha l_\theta(y_i \mid \boldsymbol{q}_i)
\end{equation}
where larger $\alpha$ means more weight on our adjustment (we set it to 0.5) and $\alpha=0$ reduces to standard negative log-likelihood. If we identify an external knowledge $\boldsymbol{M}^*$ conditionally independent to the generation, $l_{\theta}(y_i \mid \boldsymbol{M}^*, \boldsymbol{q}_i) = l_{\theta}(y_i \mid \boldsymbol{q}_i)$, even a non-zero $\alpha$ would not have an impact on the original output distribution.

\paragraph{Knowledge Transfer with Self-Matching}
Recall that during the aggregation process, we calculate the attention weights between the query and the memory bank. These weights represent the contribution of each document to answering the current question to some extent. Therefore, this inspires us to leverage the nature of the task to capture significant features of memory during continual learning. Note that the specific vector $\rvq \in \mathbb{R}^d$ for each query has the same dimension as the memory unit $\rvm_i \in \mathbb{R}^d$. Then, we calculate the cosine similarity between the query embedding and the memory vectors for the current $i^\text{th}$ document, as the document matching score $\alpha\left[:i\right] = \cos (\rvq_i, \boldsymbol{M}\left[:i\right])$. The additional training objective for updating the $i^\text{th}$ document is to maximize the cosine similarity between $\rvm_i$ and the corresponding query embedding $\rvq$. 
To prevent the model from collapsing all documents into similar vectors, we add a uniformity term to penalize excessive similarity between document embeddings, promoting diversity by encouraging larger pairwise distances between different memory vectors.

\paragraph{Inference with Top-$k$ Aggregation} 
Additionally, we employ a filtering mechanism to handle a large memory bank during downstream task inference. Let $n$ and $|D|$ be the number of output tokens for each context and the number of memory units, respectively. Then, the memory usage of $t$ cross-attention layers in the memory aggregation becomes $t \cdot \gO(|D|n^2)$. It indicates that the memory cost of the aggregation process scales linearly with the size of the memory. Unlike previous work~\cite{mac} that reduces memory consumption using hierarchical modulation aggregation, here we use a simple but effective top-$k$ filtering method. Specifically, for a given memory bank with $k$ units, we first compute the similarity between them and query, then sort by similarity and filter according to the window size used during training. Hence, the memory bank window size seen by the model in the aggregation process during training and testing is consistent, which requiring no additional training or modifications to the primary training objective. It also reduces the impact of overfitting and noise, and yields better results. Similar observations are reported in retrieval-augmented generation works~\cite{DBLP:journals/corr/abs-2306-17563,DBLP:conf/sigir/CuconasuTSFCMTS24}.

\section{Experiments}
\label{sec:exp}

In this section, we first introduce the 3 datasets used in the experiments and the details of the experimental setup, as well as the continual learning methods for comparison. We conduct extensive experiments on these Question Answering (QA) benchmark datasets to answer the following Research Questions (\textbf{RQs}):
\begin{itemize}
    \item \textbf{RQ1}: How does our model contribute to QA accuracy compared with other state-of-the-art methods?
\item \textbf{RQ2}: How effective are the key components in our model, such as the self-matching of memory vectors?
\item  \textbf{RQ3}: Can our model demonstrate robustness against knowledge interference from irrelevant documents?
\item  \textbf{RQ4}: How does our model perform in terms of the forgetting and plasticity dynamics within the document stream?
\end{itemize}


\subsection{Datasets}


Following previous works~\cite{hu2023metalearning,mac}, we employ three QA datasets, repurposing them for online adaptation to accommodate a stream of new documents.

\paragraph{StreamingQA} The StreamingQA dataset~\cite{liška2022streamingqa} features both human-written and language model-generated questions. These questions are sourced from English WMT news articles published between 2007 and 2020. Each question is linked to a complete WMT news article, with an average length of about 500 tokens per article. For learning purposes, question-article pairs from post-2018 publications are used, resulting in 21k training questions, 1.7k validation questions, and 5k test questions. 

\paragraph{SQuAD} The Stanford Question Answering Dataset (SQuAD)~\cite{rajpurkar2016squad} includes questions created by humans based on Wikipedia articles. The answer to each question is a text span from a specific paragraph within the article. Typically, paragraphs are around 150 tokens. We utilize the validation set of SQuAD as our test set and divide the training set into four additional splits. It results in 39.9k training questions, 5.6k validation questions, and 10.6k test questions. Additionally, we use 8.6k training documents, 1.2k validation documents, and 2.1k test documents.

\paragraph{ArchivalQA} The ArchivalQA dataset~\cite{wang2022archivalqa} contains questions generated automatically from the New York Times Annotated Corpus~\cite{nyt}. Each answer is a text span within an article, with questions paired to paragraphs from NYT articles. We split the validation set of ArchivalQA into five segments for our study. This setup provides us with 21.7k training questions, 5.3k validation questions, and 8.7k test questions. For documents, we utilize 12.8k training documents, 3.0k validation documents, and 5.0k test documents.

\begin{table*}[t]

\begin{center}
\small
\begin{tabular}{l l cc cc cc cc}
        \toprule
        \multirow{2}{*}{\textbf{Datasets}} & \multirow{2}{*}{\textbf{Method}} & \multicolumn{2}{c}{\textbf{DistilGPT2}} & \multicolumn{2}{c}{\textbf{GPT2-Large}} & \multicolumn{2}{c}{\textbf{GPT2-XL}} & \multicolumn{2}{c}{\textbf{Llama-2}} \\
        \cmidrule{3-10}
         & & EM & $F_1$  & EM & $F_1$  & EM  & $F_1$  & EM & $F_1$\\
        \midrule
         
        \multirow{5}{*}{\makecell{StreamingQA}} 
        & Uniform & \phantom{0}1.62 & \phantom{0}3.76  & \phantom{0}4.74 & \phantom{0}7.00 & \phantom{0}5.11 & \phantom{0}7.48  & 12.43 & 13.54  \\
        & Salient Spans  & \phantom{0}1.44 & \phantom{0}4.67 & \phantom{0}4.86 & \phantom{0}8.54 & \phantom{0}5.40 & \phantom{0}9.42  & 13.33 & 18.97 \\
        & CaMeLS  & \phantom{0}{1.62} & \phantom{0}{5.79} & \phantom{0}5.35 & 10.60 & \phantom{0}6.55 & 11.67 & - & -\\
        & MAC & \phantom{0}5.59 & 10.18 & \phantom{0}7.25 & 13.31 & \phantom{0}8.99 & 15.38 & 14.29 & 21.79 \\
        \cmidrule{2-10}
        & \textbf{CMT} (ours) & \phantom{0}\textbf{6.43} & \textbf{12.32} & \phantom{0}\textbf{7.32} & \textbf{13.43} & \phantom{0}\textbf{9.61} & \textbf{16.48} & \textbf{18.36} & \textbf{25.98} \\
        \midrule
        \multirow{5}{*}{\makecell{SQuAD}} 
        & Uniform  & \phantom{0}1.24 & \phantom{0}2.54 & \phantom{0}3.64 & \phantom{0}4.97 & \phantom{0}6.10 & \phantom{0}6.78  & 13.25 & 17.01 \\
        & Salient Spans & \phantom{0}1.03 & \phantom{0}2.47  & \phantom{0}4.03 & \phantom{0}6.48 & \phantom{0}4.55 & \phantom{0}6.74 & 13.74 & 18.66 \\
        & CaMeLS  & \phantom{0}1.47 & \phantom{0}3.08 & \phantom{0}4.97 & \phantom{0}8.63 & \phantom{0}6.70 & 10.15 & - & - \\
        & MAC & \phantom{0}2.01 & \phantom{0}6.85 & \phantom{0}6.43 & 11.42 & \phantom{0}7.10 & 12.55  & 15.07 & 21.14 \\
        \cmidrule{2-10}
        & \textbf{CMT} (ours) & \phantom{0}\textbf{3.12} & \phantom{0}\textbf{7.59} & \phantom{0}\textbf{7.15} & \textbf{12.45} & \phantom{0}\textbf{9.81} & \textbf{12.85}  & \textbf{19.54} & \textbf{25.50}\\
        \midrule
        \multirow{5}{*}{\makecell{ArchivalQA}} 
        & Uniform & \phantom{0}4.86 & \phantom{0}4.08 & \phantom{0}7.66 & \phantom{0}8.71  & \phantom{0}8.61 & 10.78  & 18.53 & 21.35 \\
        & Salient Spans & \phantom{0}4.52 & \phantom{0}3.76 & \phantom{0}9.75 & 11.19 & 11.81 & 14.11 & 18.97 & 22.75 \\
        & CaMeLS  & \phantom{0}4.62 & \phantom{0}6.19  & \phantom{0}9.92 & 12.41 & 13.87 & 15.74 & - & - \\
        & MAC   & \phantom{0}7.55 & 10.58 & 11.84 & 15.26 & 14.01 & 17.12 & 20.12 & 23.90\\
        \cmidrule{2-10}
        & \textbf{CMT} (ours)  & \phantom{0}\textbf{8.15} & \textbf{11.03} & \textbf{12.28} & \textbf{16.12} & \textbf{14.55} & \textbf{18.01} & \textbf{21.73} & \textbf{25.40} \\
        \bottomrule
       
\end{tabular}
\end{center}
\vspace{-2mm}
\caption{
Performance comparisons of online adaptation with CMT are presented. We report the Exact Match (EM) and $F_1$ scores after adapting the LLMs on a stream of documents and subsequently conducting downstream QA for test. We use the average of 3 random seeds and baseline results are from the corresponding papers. The \textbf{boldfaced} means the best results for this dataset.
}
\vspace{-4mm}
\label{tab:main}
\end{table*}

\subsection{Setup}
\paragraph{Experiment Settings} 
We conducted extensive experiments using 4 LLMs as backbones, including the GPT-2 series ~\cite{radford2018improving} and the Llama-2 series ~\cite{touvron2023llama}. The model parameters are 82M, 774M, 1.5B, and 7B, respectively. Larger models, such as the 70B, were not included due to insufficient computational resources for training. For the compressor, unlike previous work, we chose a larger decoder model, Llama-2-7b, and used Parameter-Efficient Fine-Tuning (PEFT) with a rank of 6 and a LoRA alpha set to 32, employing a different encoding method as well. We performed 1,000 steps of pre-training using English Wikipedia with auto-encoder tasks, as additional steps did not yield consistent improvements. The number of soft tokens used is 24. We evaluated online adaptation performance on a test dataset composed of documents and QA pairs. Following~\citet{mac}, we adapted the LLMs using 1,665 documents and then assessed its performance post-adaptation. To test the model's general understanding, QA pairs were sampled from these documents. Each document contains up to 1,024 tokens. Cross-attention involves 4 blocks. The batch sizes for updates and validation are 8 and 16 respectively, with gradient accumulation over 4 steps.  The optimizer is AdamW, and training will run for 50 epochs with validation every 250 steps. The learning rate is set to $1e^{-6}$ with a warmup ratio of 0.01 and a constant-with-warmup schedule. During training, we sample the document memory in the same batch and at inference $k$ is equal to training batch size. We run  the experiments on the NVIDIA A100 80G GPUs.
\paragraph{Baseline} 
We include the online fine-tuning baselines introduced in~\citet{mac}, including Uniform, Salient Spans, CaMeLS and MAC.  The uniform baseline uses uniform token weighting kearning documents and involves additional fine-tuning for question answering after adaptation. Salient Spans assigns uniform weights to tokens in salient spans~\cite{guu2020retrieval} and no weights to other tokens. CaMeLS leverages the output of a token-weighting language model (i.e., meta-learned to predict important tokens to maximize the performance of the adapted LLM).
Memory of Amortized Contexts (MAC) is an efficient online learning framework that uses the modulation to integrate new document knowledge.



\begin{table*}[h]
  \centering
   \begin{tabular}{@{}l l cc cc cc@{}}
\toprule
\multirow{2}{*}{\textbf{\#}} & \multirow{2}{*}{\textbf{Method}} & \multicolumn{2}{c}{\textbf{StreamingQA}}              & \multicolumn{2}{c}{\textbf{SQuAD}}                & \multicolumn{2}{c}{\textbf{ArchivalQA}}           \\ 
\cmidrule{3-8}
                    &                        & EM           & $F_1$                  & EM           & $F_1$             & EM           & $F_1$                 \\ \midrule
(1)                   & \textbf{CMT}              & \underline{18.36} & \textbf{25.98}          & \textbf{19.54} & \textbf{25.50}          & \underline{21.73} &    \underline{25.40}        \\
(2)                  & \quad w/o Memory-Aware Objective                   & \textbf{18.54}          & \underline{23.71}        & 15.38          & 22.77          & 20.89         &          24.18         \\
(3)                   & \quad w/o Self-Matching                   & 17.87 & 22.54        & 17.97         & 23.40       & \textbf{22.43}          & \textbf{25.68}                        \\
(4)                   & \quad w/o Top-$k$ Aggregation                   & 16.43         & 20.13         & \underline{18.35} & \underline{24.12}         & 21.09                       & 23.99         \\ \bottomrule
\end{tabular}
\vspace{-2mm}
\caption{Results of ablation study where the best results are \textbf{boldfaced} and the second-best results are \underline{underlined}. }
\vspace{-4mm}
  \label{ablation}
\end{table*}

\subsection{Model Comparison (RQ1)}
Table~\ref{tab:main} illustrates the performance of CMT in online adaptation compared to other baselines. CMT consistently outperforms these baseline methods across all datasets and models, demonstrating its superior capabilities in continual learning and knowledge retention on online adaption. These advantages are particularly evident in larger models, indicating that CMT effectively scales with model size and complexity. For instance, in the StreamingQA dataset, CMT consistently surpasses other methods for all model variants. Specifically, with DistilGPT2, CMT achieves an EM score of 6.43 and an $F_1$ score of 12.32, outperforming the next best method, MAC, which scores 5.59 (EM) and 10.18 ($F_1$). This performance gap widens with larger models, with CMT achieving the highest scores on Llama-2 (EM: 18.36, $F_1$: 25.98). This demonstrates CMT's superior ability to incorporate and retain new knowledge in real-time. Furthermore, CMT is efficient in terms of memory usage and adaptation time. Unlike CaMeLS, CMT does not require gradient computation for updates. Furthermore, CMT reduces the proportion of trainable parameters by 5.4 times and inference time due to top-$k$ aggregation compared to MAC, facilitating easier scalability with larger document corpora and model sizes.

\begin{figure}[!]
\centering
\resizebox{0.9\linewidth}{!}{
\includegraphics[]{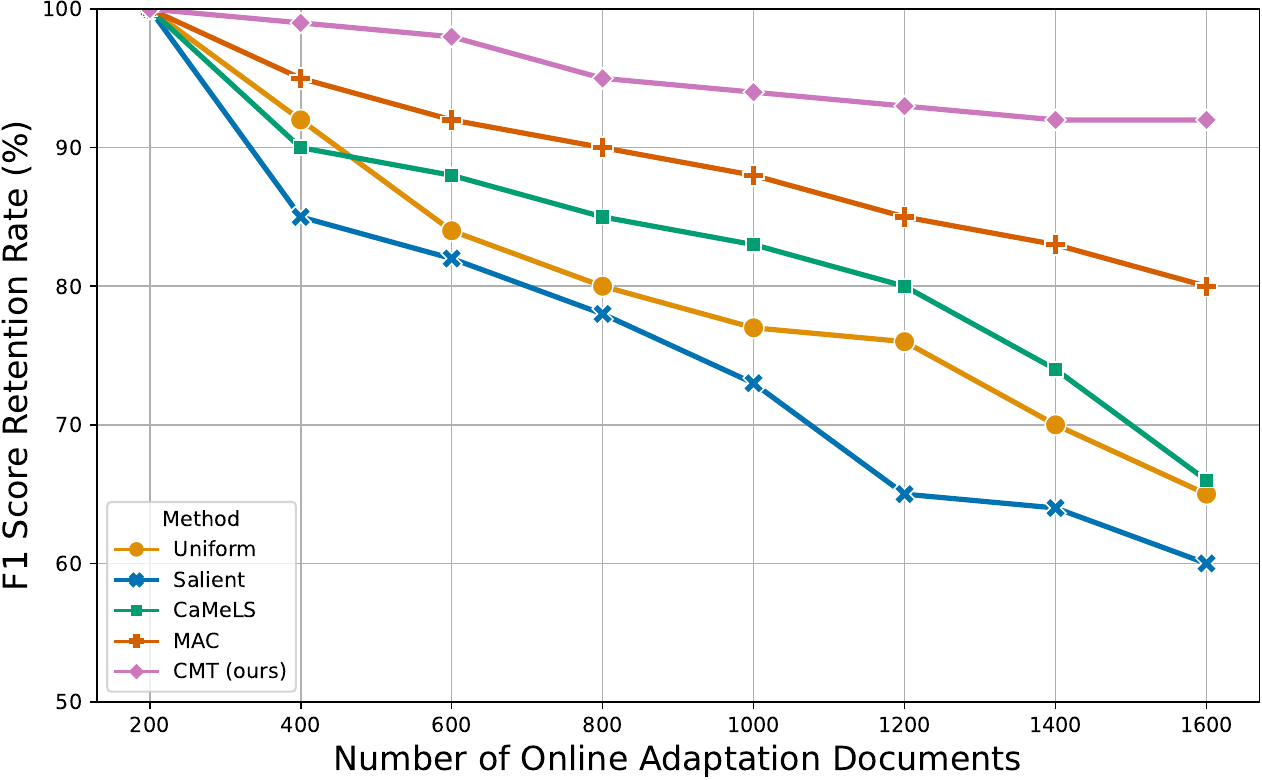}
}
\caption{Knowledge retention analysis under~\texttt{Llama-2-7b} trained on StreamingQA dataset.}
\vspace{-2mm}
\label{fig:retention}
\end{figure}

\subsection{Ablation Study (RQ2)}
This section presents an ablation study to assess the impact of various components of the CMT on its performance. Table~\ref{ablation} summarizes the results across three datasets. We evaluate the full CMT model (1) and three variants: CMT without the Memory-Aware Objective (2), CMT without Self-Matching (3), and CMT without Top-$k$ Aggregation (4). For example, in the StreamingQA dataset, the full CMT model achieves an EM score of 18.36 and an $F_1$ score of 25.98, outperforming all other variants. The removal of the Memory-Aware Objective (variant 2) slightly increases the EM score to 18.54 but decreases the $F_1$ score to 23.71. The absence of Self-Matching (variant 3) results in lower scores (EM: 17.87, $F_1$: 22.54), indicating the importance of this component. The variant without Top-$k$ Aggregation (4) shows the lowest performance (EM: 16.43, $F_1$: 20.13), highlighting its critical role in CMT.
The ablation study reveals the relative importance of each CMT component. The Memory-Aware Objective and Top-$k$ Aggregation are crucial for maximizing performance across most datasets. Self-Matching, while generally beneficial, can sometimes be omitted without severe performance degradation, as seen in the ArchivalQA results. However, the full CMT model consistently provides the best or near-best performance, validating the integrated approach.
\begin{figure}[!]
\centering
\resizebox{0.9\linewidth}{!}{
\includegraphics[]{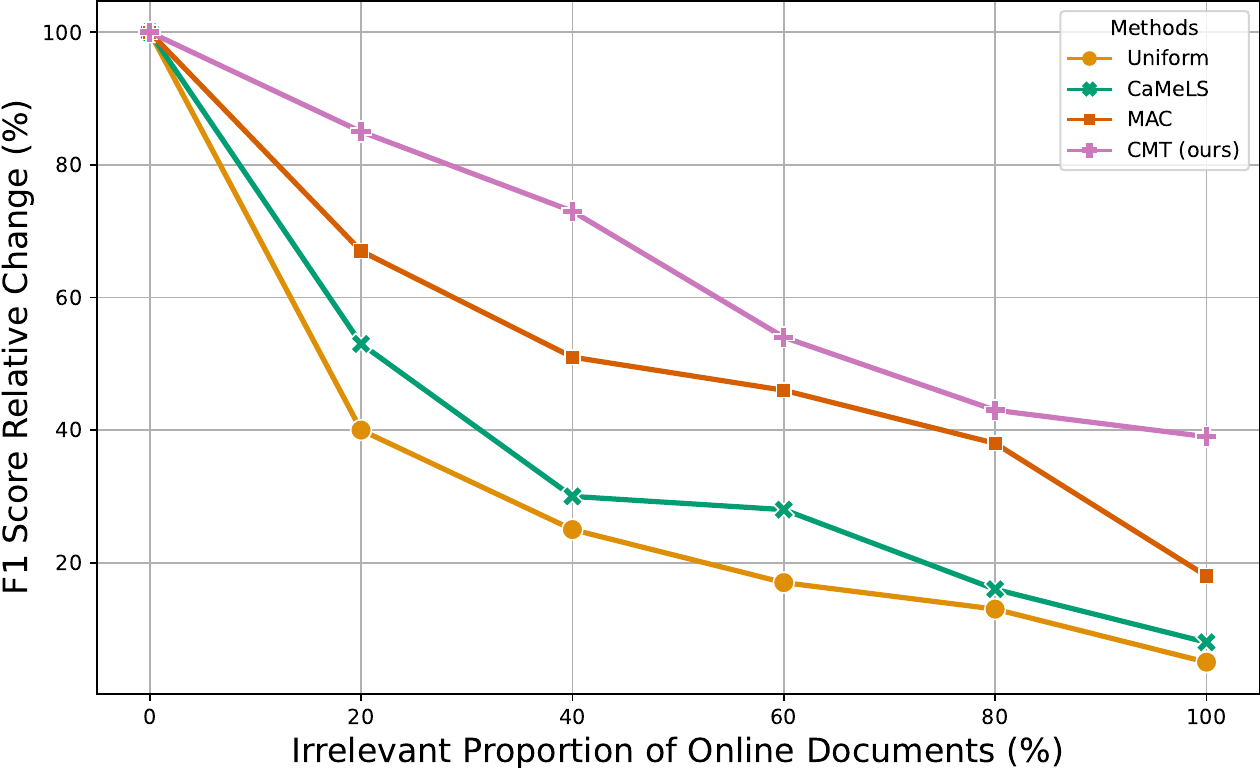}
}
\caption{Performance of robustness analysis experiments. We use synthetic unrelated documents to test the impact of irrelevant interference brought by memory integration.}
\vspace{-4mm}
\label{fig:robust}
\end{figure}
\subsection{Knowledge Retention (RQ3)}
Following~\citet{mac}, we evaluate retention ratio determined by the decline in the $F_1$ score of the initially adapted 200 documents during online adaption. As shown in Figure~\ref{fig:retention}, CMT and CaMeLS lead in performance, followed by MAC, Salient, and finally the Uniform method. The plot indicates that all methods reduce from an increased number of online adaptation documents, as shown by the downward trends in both $F_1$ score and retention rate. However, the rate of improvement varies among the methods.
Methods like CMT and CaMeLS show a higher scalability factor, indicating that they are better suited for environments where the volume of adaptation data is substantial. The gap between the highest (CMT) and the lowest performing method (Uniform) widens as the number of documents increases, highlighting the importance of choosing a more efficient method for large-scale online adaptation tasks.

\subsection{Robustness Analysis (RQ4)} 

We make use of irrelevant synthetic data, which is obtained by generating text that is irrelevant to the current document using \texttt{gpt-4o}. As shown in Figure~\ref{fig:robust}, the performance of the Uniform shows a significant decline as the proportion of irrelevant documents increases. Initially maintaining a high relative $F_1$ score, the performance deteriorates sharply beyond the 20\%, indicating a high sensitivity to irrelevant data. CaMeLS exhibits a more robust performance compared to the Uniform method. However, a noticeable performance drop is still observed beyond the 60\% threshold. The MAC method demonstrates a relatively stable performance across different proportions of irrelevant documents. While there is a gradual decline in the $F_1$ score, it is less pronounced compared to the Uniform and CaMeLS methods, highlighting MAC's effectiveness in handling irrelevant data. Among the four methods, CMT shows the best performance stability. The relative change in $F_1$ score remains minimal even as the proportion of irrelevant documents approaches 100\%. This indicates that CMT is highly robust to irrelevant data, maintaining high accuracy and reliability. 
\section{Conclusion}
In this paper, we propose a continual learning method for LLMs named CMT, which consists of the memory bank in a latent space serving as the model's updatable knowledge parameters. CMT can update the memory with new knowledge, enabling effective knowledge integration and slow forgetting of prior knowledge. Comparisons with baselines on three datasets, along with evaluations of knowledge retention and robustness, demonstrate the advantages of CMT in effectively absorbing new knowledge and retaining knowledge. In the future,  we plan to further leverage the memory mechanism of the LLMs itself for continual learning.

\bibliography{aaai25}

\begin{thebibliography}{65}
\providecommand{\natexlab}[1]{#1}

\bibitem[{Aljundi et~al.(2018)Aljundi, Babiloni, Elhoseiny, Rohrbach, and Tuytelaars}]{aljundi2018memory}
Aljundi, R.; Babiloni, F.; Elhoseiny, M.; Rohrbach, M.; and Tuytelaars, T. 2018.
\newblock Memory aware synapses: Learning what (not) to forget.
\newblock In \emph{Proceedings of the European conference on computer vision (ECCV)}, 139--154.

\bibitem[{Anil et~al.(2023)Anil, Borgeaud, Wu, Alayrac, Yu, Soricut, Schalkwyk, Dai, Hauth, Millican, Silver, Petrov, Johnson, Antonoglou, Schrittwieser, Glaese, Chen, Pitler, Lillicrap, Lazaridou, Firat, Molloy, Isard, Barham, Hennigan, Lee, Viola, Reynolds, Xu, Doherty, Collins, Meyer, Rutherford, Moreira, Ayoub, Goel, Tucker, Piqueras, Krikun, Barr, Savinov, Danihelka, Roelofs, White, Andreassen, von Glehn, Yagati, Kazemi, Gonzalez, Khalman, Sygnowski, and et~al.}]{DBLP:journals/corr/abs-2312-11805}
Anil, R.; Borgeaud, S.; Wu, Y.; Alayrac, J.; Yu, J.; Soricut, R.; Schalkwyk, J.; Dai, A.~M.; Hauth, A.; Millican, K.; Silver, D.; Petrov, S.; Johnson, M.; Antonoglou, I.; Schrittwieser, J.; Glaese, A.; Chen, J.; Pitler, E.; Lillicrap, T.~P.; Lazaridou, A.; Firat, O.; Molloy, J.; Isard, M.; Barham, P.~R.; Hennigan, T.; Lee, B.; Viola, F.; Reynolds, M.; Xu, Y.; Doherty, R.; Collins, E.; Meyer, C.; Rutherford, E.; Moreira, E.; Ayoub, K.; Goel, M.; Tucker, G.; Piqueras, E.; Krikun, M.; Barr, I.; Savinov, N.; Danihelka, I.; Roelofs, B.; White, A.; Andreassen, A.; von Glehn, T.; Yagati, L.; Kazemi, M.; Gonzalez, L.; Khalman, M.; Sygnowski, J.; and et~al. 2023.
\newblock Gemini: {A} Family of Highly Capable Multimodal Models.
\newblock \emph{CoRR}, abs/2312.11805.

\bibitem[{Ba et~al.(2016)Ba, Hinton, Mnih, Leibo, and Ionescu}]{ba2016using}
Ba, J.; Hinton, G.; Mnih, V.; Leibo, J.~Z.; and Ionescu, C. 2016.
\newblock Using fast weights to attend to the recent past.
\newblock In \emph{Proceedings of the 30th International Conference on Neural Information Processing Systems}, 4338--4346.

\bibitem[{Bjork(1994)}]{bjork1994memory}
Bjork, R.~A. 1994.
\newblock Memory and Metamemory Considerations in the.
\newblock \emph{Metacognition: Knowing about Knowing}, 185.

\bibitem[{Brown et~al.(2020)Brown, Mann, Ryder, Subbiah, Kaplan, Dhariwal, Neelakantan, Shyam, Sastry, Askell, Agarwal, Herbert{-}Voss, Krueger, Henighan, Child, Ramesh, Ziegler, Wu, Winter, Hesse, Chen, Sigler, Litwin, Gray, Chess, Clark, Berner, McCandlish, Radford, Sutskever, and Amodei}]{DBLP:conf/nips/BrownMRSKDNSSAA20}
Brown, T.~B.; Mann, B.; Ryder, N.; Subbiah, M.; Kaplan, J.; Dhariwal, P.; Neelakantan, A.; Shyam, P.; Sastry, G.; Askell, A.; Agarwal, S.; Herbert{-}Voss, A.; Krueger, G.; Henighan, T.; Child, R.; Ramesh, A.; Ziegler, D.~M.; Wu, J.; Winter, C.; Hesse, C.; Chen, M.; Sigler, E.; Litwin, M.; Gray, S.; Chess, B.; Clark, J.; Berner, C.; McCandlish, S.; Radford, A.; Sutskever, I.; and Amodei, D. 2020.
\newblock Language Models are Few-Shot Learners.
\newblock In Larochelle, H.; Ranzato, M.; Hadsell, R.; Balcan, M.; and Lin, H., eds., \emph{Advances in Neural Information Processing Systems 33: Annual Conference on Neural Information Processing Systems 2020, NeurIPS 2020, December 6-12, 2020, virtual}.

\bibitem[{Bulatov, Kuratov, and Burtsev(2022)}]{RMT}
Bulatov, A.; Kuratov, Y.; and Burtsev, M.~S. 2022.
\newblock Recurrent Memory Transformer.
\newblock In \emph{NeurIPS}.

\bibitem[{Burtsev and Sapunov(2020)}]{memoryTransformer}
Burtsev, M.~S.; and Sapunov, G.~V. 2020.
\newblock Memory Transformer.
\newblock \emph{CoRR}, abs/2006.11527.

\bibitem[{Buzzega et~al.(2020)Buzzega, Boschini, Porrello, Abati, and Calderara}]{buzzega2020dark}
Buzzega, P.; Boschini, M.; Porrello, A.; Abati, D.; and Calderara, S. 2020.
\newblock Dark experience for general continual learning: a strong, simple baseline.
\newblock \emph{Advances in neural information processing systems}, 33: 15920--15930.

\bibitem[{Cheng et~al.(2024)Cheng, Wang, Zhang, Ge, Chen, Wei, Zhang, and Zhao}]{xrag}
Cheng, X.; Wang, X.; Zhang, X.; Ge, T.; Chen, S.; Wei, F.; Zhang, H.; and Zhao, D. 2024.
\newblock xRAG: Extreme Context Compression for Retrieval-augmented Generation with One Token.
\newblock \emph{CoRR}, abs/2405.13792.

\bibitem[{Chevalier et~al.(2023)Chevalier, Wettig, Ajith, and Chen}]{chevalier2023adapting}
Chevalier, A.; Wettig, A.; Ajith, A.; and Chen, D. 2023.
\newblock Adapting Language Models to Compress Contexts.
\newblock In \emph{Proceedings of the 2023 Conference on Empirical Methods in Natural Language Processing}, 3829--3846.

\bibitem[{Cuconasu et~al.(2024)Cuconasu, Trappolini, Siciliano, Filice, Campagnano, Maarek, Tonellotto, and Silvestri}]{DBLP:conf/sigir/CuconasuTSFCMTS24}
Cuconasu, F.; Trappolini, G.; Siciliano, F.; Filice, S.; Campagnano, C.; Maarek, Y.; Tonellotto, N.; and Silvestri, F. 2024.
\newblock The Power of Noise: Redefining Retrieval for {RAG} Systems.
\newblock In Yang, G.~H.; Wang, H.; Han, S.; Hauff, C.; Zuccon, G.; and Zhang, Y., eds., \emph{Proceedings of the 47th International {ACM} {SIGIR} Conference on Research and Development in Information Retrieval, {SIGIR} 2024, Washington DC, USA, July 14-18, 2024}, 719--729. {ACM}.

\bibitem[{Ge et~al.(2024)Ge, Jing, Wang, Wang, Chen, and Wei}]{gecontext}
Ge, T.; Jing, H.; Wang, L.; Wang, X.; Chen, S.-Q.; and Wei, F. 2024.
\newblock In-context Autoencoder for Context Compression in a Large Language Model.
\newblock In \emph{The Twelfth International Conference on Learning Representations}.

\bibitem[{Guu et~al.(2020)Guu, Lee, Tung, Pasupat, and Chang}]{guu2020retrieval}
Guu, K.; Lee, K.; Tung, Z.; Pasupat, P.; and Chang, M. 2020.
\newblock Retrieval augmented language model pre-training.

\bibitem[{He et~al.(2024)He, Qin, Prakriya, Sun, and Cong}]{hmt}
He, Z.; Qin, Z.; Prakriya, N.; Sun, Y.; and Cong, J. 2024.
\newblock {HMT:} Hierarchical Memory Transformer for Long Context Language Processing.
\newblock \emph{CoRR}, abs/2405.06067.

\bibitem[{Hu et~al.(2023)Hu, Mitchell, Manning, and Finn}]{hu2023metalearning}
Hu, N.; Mitchell, E.; Manning, C.~D.; and Finn, C. 2023.
\newblock Meta-Learning Online Adaptation of Language Models.

\bibitem[{Karpicke and Roediger~III(2008)}]{karpicke2008critical}
Karpicke, J.~D.; and Roediger~III, H.~L. 2008.
\newblock The critical importance of retrieval for learning.
\newblock \emph{science}, 319(5865): 966--968.

\bibitem[{Khandelwal et~al.(2019)Khandelwal, Levy, Jurafsky, Zettlemoyer, and Lewis}]{kNNLM}
Khandelwal, U.; Levy, O.; Jurafsky, D.; Zettlemoyer, L.; and Lewis, M. 2019.
\newblock Generalization through memorization: Nearest neighbor language models.
\newblock \emph{arXiv preprint arXiv:1911.00172}.

\bibitem[{Kim et~al.(2019)Kim, Mnih, Schwarz, Garnelo, Eslami, Rosenbaum, Vinyals, and Teh}]{kim2019attentive}
Kim, H.; Mnih, A.; Schwarz, J.; Garnelo, M.; Eslami, A.; Rosenbaum, D.; Vinyals, O.; and Teh, Y.~W. 2019.
\newblock Attentive neural processes.

\bibitem[{Kim et~al.(2023)Kim, Yeom, Yun, and Song}]{DBLP:journals/corr/abs-2312-03414}
Kim, J.; Yeom, J.; Yun, S.; and Song, H.~O. 2023.
\newblock Compressed Context Memory For Online Language Model Interaction.
\newblock \emph{CoRR}, abs/2312.03414.

\bibitem[{Kirkpatrick et~al.(2017)Kirkpatrick, Pascanu, Rabinowitz, Veness, Desjardins, Rusu, Milan, Quan, Ramalho, Grabska-Barwinska et~al.}]{kirkpatrick2017overcoming}
Kirkpatrick, J.; Pascanu, R.; Rabinowitz, N.; Veness, J.; Desjardins, G.; Rusu, A.~A.; Milan, K.; Quan, J.; Ramalho, T.; Grabska-Barwinska, A.; et~al. 2017.
\newblock Overcoming catastrophic forgetting in neural networks.
\newblock \emph{Proceedings of the national academy of sciences}, 114(13): 3521--3526.

\bibitem[{Li and Hoiem(2017)}]{li2017learning}
Li, Z.; and Hoiem, D. 2017.
\newblock Learning without forgetting.
\newblock \emph{IEEE transactions on pattern analysis and machine intelligence}, 40(12): 2935--2947.

\bibitem[{Liška et~al.(2022)Liška, Kočiský, Gribovskaya, Terzi, Sezener, Agrawal, de~Masson~d'Autume, Scholtes, Zaheer, Young, Gilsenan-McMahon, Austin, Blunsom, and Lazaridou}]{liška2022streamingqa}
Liška, A.; Kočiský, T.; Gribovskaya, E.; Terzi, T.; Sezener, E.; Agrawal, D.; de~Masson~d'Autume, C.; Scholtes, T.; Zaheer, M.; Young, S.; Gilsenan-McMahon, E.; Austin, S.; Blunsom, P.; and Lazaridou, A. 2022.
\newblock StreamingQA: A Benchmark for Adaptation to New Knowledge over Time in Question Answering Models.
\newblock arXiv:2205.11388.

\bibitem[{Llama~Team(2024)}]{2023-Llama2}
Llama~Team, A. .~M. 2024.
\newblock The Llama 3 Herd of Models.

\bibitem[{McCloskey and Cohen(1989)}]{mccloskey1989catastrophic}
McCloskey, M.; and Cohen, N.~J. 1989.
\newblock Catastrophic Interference in Connectionist Networks: The Sequential Learning Problem.
\newblock volume~24 of \emph{Psychology of Learning and Motivation}, 109--165. Academic Press.

\bibitem[{Modarressi et~al.(2023)Modarressi, Imani, Fayyaz, and Sch{\"u}tze}]{ret-llm}
Modarressi, A.; Imani, A.; Fayyaz, M.; and Sch{\"u}tze, H. 2023.
\newblock RET-LLM: Towards a General Read-Write Memory for Large Language Models.
\newblock \emph{arXiv preprint arXiv:2305.14322}.

\bibitem[{Moro et~al.(2023)Moro, Ragazzi, Valgimigli, Frisoni, Sartori, and Marfia}]{Memory-Enhanced-Transformer}
Moro, G.; Ragazzi, L.; Valgimigli, L.; Frisoni, G.; Sartori, C.; and Marfia, G. 2023.
\newblock Efficient Memory-Enhanced Transformer for Long-Document Summarization in Low-Resource Regimes.
\newblock \emph{Sensors}, 23(7): 3542.

\bibitem[{Mu, Li, and Goodman(2024)}]{mu2024learning}
Mu, J.; Li, X.; and Goodman, N. 2024.
\newblock Learning to compress prompts with gist tokens.
\newblock \emph{Advances in Neural Information Processing Systems}, 36.

\bibitem[{OpenAI(2023)}]{DBLP:journals/corr/abs-2303-08774}
OpenAI. 2023.
\newblock {GPT-4} Technical Report.
\newblock \emph{CoRR}, abs/2303.08774.

\bibitem[{Phang et~al.(2023)Phang, Mao, He, and Chen}]{phang2023hypertuning}
Phang, J.; Mao, Y.; He, P.; and Chen, W. 2023.
\newblock Hypertuning: Toward adapting large language models without back-propagation.
\newblock In \emph{International Conference on Machine Learning}, 27854--27875. PMLR.

\bibitem[{Pyc and Rawson(2010)}]{pyc2010testing}
Pyc, M.~A.; and Rawson, K.~A. 2010.
\newblock Why testing improves memory: Mediator effectiveness hypothesis.
\newblock \emph{Science}, 330(6002): 335--335.

\bibitem[{Qin et~al.(2023)Qin, Jagerman, Hui, Zhuang, Wu, Shen, Liu, Liu, Metzler, Wang, and Bendersky}]{DBLP:journals/corr/abs-2306-17563}
Qin, Z.; Jagerman, R.; Hui, K.; Zhuang, H.; Wu, J.; Shen, J.; Liu, T.; Liu, J.; Metzler, D.; Wang, X.; and Bendersky, M. 2023.
\newblock Large Language Models are Effective Text Rankers with Pairwise Ranking Prompting.
\newblock \emph{CoRR}, abs/2306.17563.

\bibitem[{Radford et~al.(2018)Radford, Narasimhan, Salimans, Sutskever et~al.}]{radford2018improving}
Radford, A.; Narasimhan, K.; Salimans, T.; Sutskever, I.; et~al. 2018.
\newblock Improving language understanding by generative pre-training.
\newblock In \emph{preprint}.

\bibitem[{Rajpurkar et~al.(2016)Rajpurkar, Zhang, Lopyrev, and Liang}]{rajpurkar2016squad}
Rajpurkar, P.; Zhang, J.; Lopyrev, K.; and Liang, P. 2016.
\newblock SQuAD: 100,000+ Questions for Machine Comprehension of Text.
\newblock arXiv:1606.05250.

\bibitem[{Riemer et~al.(2018)Riemer, Cases, Ajemian, Liu, Rish, Tu, and Tesauro}]{riemer2018learning}
Riemer, M.; Cases, I.; Ajemian, R.; Liu, M.; Rish, I.; Tu, Y.; and Tesauro, G. 2018.
\newblock Learning to learn without forgetting by maximizing transfer and minimizing interference.
\newblock \emph{arXiv preprint arXiv:1810.11910}.

\bibitem[{Ritter, Botev, and Barber(2018)}]{ritter2018online}
Ritter, H.; Botev, A.; and Barber, D. 2018.
\newblock Online structured laplace approximations for overcoming catastrophic forgetting.
\newblock \emph{Advances in Neural Information Processing Systems}, 31.

\bibitem[{{Sandhaus, Evan}(2008)}]{nyt}
{Sandhaus, Evan}. 2008.
\newblock The New York Times Annotated Corpus.

\bibitem[{Schwarz et~al.(2018)Schwarz, Czarnecki, Luketina, Grabska-Barwinska, Teh, Pascanu, and Hadsell}]{schwarz2018progress}
Schwarz, J.; Czarnecki, W.; Luketina, J.; Grabska-Barwinska, A.; Teh, Y.~W.; Pascanu, R.; and Hadsell, R. 2018.
\newblock Progress \& compress: A scalable framework for continual learning.
\newblock In \emph{International conference on machine learning}, 4528--4537. PMLR.

\bibitem[{Shazeer et~al.(2017)Shazeer, Mirhoseini, Maziarz, Davis, Le, Hinton, and Dean}]{moe}
Shazeer, N.; Mirhoseini, A.; Maziarz, K.; Davis, A.; Le, Q.~V.; Hinton, G.~E.; and Dean, J. 2017.
\newblock Outrageously Large Neural Networks: The Sparsely-Gated Mixture-of-Experts Layer.
\newblock In \emph{5th International Conference on Learning Representations, {ICLR} 2017, Toulon, France, April 24-26, 2017, Conference Track Proceedings}. OpenReview.net.

\bibitem[{Shi and Wang(2024)}]{shi2024unified}
Shi, H.; and Wang, H. 2024.
\newblock A Unified Approach to Domain Incremental Learning with Memory: Theory and Algorithm.
\newblock \emph{Advances in Neural Information Processing Systems}, 36.

\bibitem[{Shi et~al.(2024)Shi, Xu, Wang, Qin, Wang, Wang, and Wang}]{DBLP:journals/corr/abs-2404-16789}
Shi, H.; Xu, Z.; Wang, H.; Qin, W.; Wang, W.; Wang, Y.; and Wang, H. 2024.
\newblock Continual Learning of Large Language Models: {A} Comprehensive Survey.
\newblock \emph{CoRR}, abs/2404.16789.

\bibitem[{Smith et~al.(2023)Smith, Tian, Halbe, Hsu, and Kira}]{smith2023closer}
Smith, J.~S.; Tian, J.; Halbe, S.; Hsu, Y.-C.; and Kira, Z. 2023.
\newblock A Closer Look at Rehearsal-Free Continual Learning.
\newblock arXiv:2203.17269.

\bibitem[{Snell, Klein, and Zhong(2022)}]{snell2022learning}
Snell, C.; Klein, D.; and Zhong, R. 2022.
\newblock Learning by distilling context.
\newblock \emph{arXiv preprint arXiv:2209.15189}.

\bibitem[{Song et~al.(2023)Song, Han, Zeng, Li, Chen, Liu, Sun, and Yang}]{DBLP:journals/corr/abs-2309-14763}
Song, C.; Han, X.; Zeng, Z.; Li, K.; Chen, C.; Liu, Z.; Sun, M.; and Yang, T. 2023.
\newblock ConPET: Continual Parameter-Efficient Tuning for Large Language Models.
\newblock \emph{CoRR}, abs/2309.14763.

\bibitem[{Sprechmann et~al.(2018)Sprechmann, Jayakumar, Rae, Pritzel, Badia, Uria, Vinyals, Hassabis, Pascanu, and Blundell}]{sprechmann2018memory}
Sprechmann, P.; Jayakumar, S.~M.; Rae, J.~W.; Pritzel, A.; Badia, A.~P.; Uria, B.; Vinyals, O.; Hassabis, D.; Pascanu, R.; and Blundell, C. 2018.
\newblock Memory-based Parameter Adaptation.
\newblock In \emph{International Conference on Learning Representations}.

\bibitem[{Su et~al.(2024)Su, Ahmed, Lu, Pan, Bo, and Liu}]{ROPE}
Su, J.; Ahmed, M. H.~M.; Lu, Y.; Pan, S.; Bo, W.; and Liu, Y. 2024.
\newblock RoFormer: Enhanced transformer with Rotary Position Embedding.
\newblock \emph{Neurocomputing}, 568: 127063.

\bibitem[{Sukhbaatar et~al.(2015)Sukhbaatar, Weston, Fergus et~al.}]{E2EMN}
Sukhbaatar, S.; Weston, J.; Fergus, R.; et~al. 2015.
\newblock End-to-end memory networks.
\newblock \emph{Advances in neural information processing systems}, 28.

\bibitem[{Tack et~al.(2024)Tack, Kim, Mitchell, Shin, Teh, and Schwarz}]{mac}
Tack, J.; Kim, J.; Mitchell, E.; Shin, J.; Teh, Y.~W.; and Schwarz, J.~R. 2024.
\newblock Online Adaptation of Language Models with a Memory of Amortized Contexts.
\newblock \emph{CoRR}, abs/2403.04317.

\bibitem[{Touvron et~al.(2023{\natexlab{a}})Touvron, Martin, Stone, Albert, Almahairi, Babaei, Bashlykov, Batra, Bhargava, Bhosale, Bikel, Blecher, Canton{-}Ferrer, Chen, Cucurull, Esiobu, Fernandes, Fu, Fu, Fuller, Gao, Goswami, Goyal, Hartshorn, Hosseini, Hou, Inan, Kardas, Kerkez, Khabsa, Kloumann, Korenev, Koura, Lachaux, Lavril, Lee, Liskovich, Lu, Mao, Martinet, Mihaylov, Mishra, Molybog, Nie, Poulton, Reizenstein, Rungta, Saladi, Schelten, Silva, Smith, Subramanian, Tan, Tang, Taylor, Williams, Kuan, Xu, Yan, Zarov, Zhang, Fan, Kambadur, Narang, Rodriguez, Stojnic, Edunov, and Scialom}]{DBLP:journals/corr/abs-2307-09288}
Touvron, H.; Martin, L.; Stone, K.; Albert, P.; Almahairi, A.; Babaei, Y.; Bashlykov, N.; Batra, S.; Bhargava, P.; Bhosale, S.; Bikel, D.; Blecher, L.; Canton{-}Ferrer, C.; Chen, M.; Cucurull, G.; Esiobu, D.; Fernandes, J.; Fu, J.; Fu, W.; Fuller, B.; Gao, C.; Goswami, V.; Goyal, N.; Hartshorn, A.; Hosseini, S.; Hou, R.; Inan, H.; Kardas, M.; Kerkez, V.; Khabsa, M.; Kloumann, I.; Korenev, A.; Koura, P.~S.; Lachaux, M.; Lavril, T.; Lee, J.; Liskovich, D.; Lu, Y.; Mao, Y.; Martinet, X.; Mihaylov, T.; Mishra, P.; Molybog, I.; Nie, Y.; Poulton, A.; Reizenstein, J.; Rungta, R.; Saladi, K.; Schelten, A.; Silva, R.; Smith, E.~M.; Subramanian, R.; Tan, X.~E.; Tang, B.; Taylor, R.; Williams, A.; Kuan, J.~X.; Xu, P.; Yan, Z.; Zarov, I.; Zhang, Y.; Fan, A.; Kambadur, M.; Narang, S.; Rodriguez, A.; Stojnic, R.; Edunov, S.; and Scialom, T. 2023{\natexlab{a}}.
\newblock Llama 2: Open Foundation and Fine-Tuned Chat Models.
\newblock \emph{CoRR}, abs/2307.09288.

\bibitem[{Touvron et~al.(2023{\natexlab{b}})Touvron, Martin, Stone, Albert, Almahairi, Babaei, Bashlykov, Batra, Bhargava, Bhosale et~al.}]{touvron2023llama}
Touvron, H.; Martin, L.; Stone, K.; Albert, P.; Almahairi, A.; Babaei, Y.; Bashlykov, N.; Batra, S.; Bhargava, P.; Bhosale, S.; et~al. 2023{\natexlab{b}}.
\newblock Llama 2: Open foundation and fine-tuned chat models.
\newblock \emph{arXiv preprint arXiv:2307.09288}.

\bibitem[{Vaswani et~al.(2017)Vaswani, Shazeer, Parmar, Uszkoreit, Jones, Gomez, Kaiser, and Polosukhin}]{vaswani2017attention}
Vaswani, A.; Shazeer, N.; Parmar, N.; Uszkoreit, J.; Jones, L.; Gomez, A.~N.; Kaiser, {\L}.; and Polosukhin, I. 2017.
\newblock Attention is all you need.

\bibitem[{Wang, Jatowt, and Yoshikawa(2022)}]{wang2022archivalqa}
Wang, J.; Jatowt, A.; and Yoshikawa, M. 2022.
\newblock ArchivalQA: A Large-scale Benchmark Dataset for Open Domain Question Answering over Historical News Collections.
\newblock arXiv:2109.03438.

\bibitem[{Wang et~al.(2022)Wang, Zhang, Li, Zhu, and Zhong}]{wang2022coscl}
Wang, L.; Zhang, X.; Li, Q.; Zhu, J.; and Zhong, Y. 2022.
\newblock CoSCL: Cooperation of Small Continual Learners is Stronger Than a Big One.
\newblock In \emph{Computer Vision--ECCV 2022: 17th European Conference, Tel Aviv, Israel, October 23--27, 2022, Proceedings, Part XXVI}, 254--271. Springer.

\bibitem[{Wang et~al.(2024{\natexlab{a}})Wang, Zhang, Su, and Zhu}]{DBLP:journals/pami/WangZSZ24}
Wang, L.; Zhang, X.; Su, H.; and Zhu, J. 2024{\natexlab{a}}.
\newblock A Comprehensive Survey of Continual Learning: Theory, Method and Application.
\newblock \emph{{IEEE} Trans. Pattern Anal. Mach. Intell.}, 46(8): 5362--5383.

\bibitem[{Wang et~al.(2023)Wang, Dong, Cheng, Liu, Yan, Gao, and Wei}]{LongMEM}
Wang, W.; Dong, L.; Cheng, H.; Liu, X.; Yan, X.; Gao, J.; and Wei, F. 2023.
\newblock Augmenting Language Models with Long-Term Memory.
\newblock \emph{arXiv preprint arXiv:2306.07174}.

\bibitem[{Wang et~al.(2024{\natexlab{b}})Wang, Chen, Shang, and McAuley}]{memoryllm}
Wang, Y.; Chen, X.; Shang, J.; and McAuley, J.~J. 2024{\natexlab{b}}.
\newblock {MEMORYLLM:} Towards Self-Updatable Large Language Models.
\newblock \emph{CoRR}, abs/2402.04624.

\bibitem[{Weston, Chopra, and Bordes(2015)}]{weston2015memory}
Weston, J.; Chopra, S.; and Bordes, A. 2015.
\newblock Memory networks.
\newblock In \emph{3rd International Conference on Learning Representations, ICLR 2015}.

\bibitem[{Wingate, Shoeybi, and Sorensen(2022)}]{wingate2022prompt}
Wingate, D.; Shoeybi, M.; and Sorensen, T. 2022.
\newblock Prompt Compression and Contrastive Conditioning for Controllability and Toxicity Reduction in Language Models.
\newblock In \emph{Findings of the Association for Computational Linguistics: EMNLP 2022}, 5621--5634.

\bibitem[{Wortsman et~al.(2022)Wortsman, Ilharco, Gadre, Roelofs, Gontijo-Lopes, Morcos, Namkoong, Farhadi, Carmon, Kornblith et~al.}]{wortsman2022model}
Wortsman, M.; Ilharco, G.; Gadre, S.~Y.; Roelofs, R.; Gontijo-Lopes, R.; Morcos, A.~S.; Namkoong, H.; Farhadi, A.; Carmon, Y.; Kornblith, S.; et~al. 2022.
\newblock Model soups: averaging weights of multiple fine-tuned models improves accuracy without increasing inference time.

\bibitem[{Wu et~al.(2024)Wu, Luo, Li, Pan, Vu, and Haffari}]{DBLP:journals/corr/abs-2402-01364}
Wu, T.; Luo, L.; Li, Y.; Pan, S.; Vu, T.; and Haffari, G. 2024.
\newblock Continual Learning for Large Language Models: {A} Survey.
\newblock \emph{CoRR}, abs/2402.01364.

\bibitem[{Xu et~al.(2020)Xu, Ton, Kim, Kosiorek, and Teh}]{xu2020metafun}
Xu, J.; Ton, J.-F.; Kim, H.; Kosiorek, A.~R.; and Teh, Y.~W. 2020.
\newblock MetaFun: Meta-Learning with Iterative Functional Updates.

\bibitem[{Yang et~al.(2024)Yang, Lin, Wang, Wu, Li, Tang, Wei, Wang, Tang, Song et~al.}]{yang2024text}
Yang, H.; Lin, Z.; Wang, W.; Wu, H.; Li, Z.; Tang, B.; Wei, W.; Wang, J.; Tang, Z.; Song, S.; et~al. 2024.
\newblock $\text{Memory}^3$: Language Modeling with Explicit Memory.
\newblock \emph{arXiv preprint arXiv:2407.01178}.

\bibitem[{Yao et~al.(2023)Yao, Wang, Tian, Cheng, Li, Deng, Chen, and Zhang}]{DBLP:conf/emnlp/YaoWT0LDC023}
Yao, Y.; Wang, P.; Tian, B.; Cheng, S.; Li, Z.; Deng, S.; Chen, H.; and Zhang, N. 2023.
\newblock Editing Large Language Models: Problems, Methods, and Opportunities.
\newblock In Bouamor, H.; Pino, J.; and Bali, K., eds., \emph{Proceedings of the 2023 Conference on Empirical Methods in Natural Language Processing, {EMNLP} 2023, Singapore, December 6-10, 2023}, 10222--10240. Association for Computational Linguistics.

\bibitem[{Zheng et~al.(2024)Zheng, Qiu, Shi, and Ma}]{DBLP:journals/corr/abs-2406-06391}
Zheng, J.; Qiu, S.; Shi, C.; and Ma, Q. 2024.
\newblock Towards Lifelong Learning of Large Language Models: {A} Survey.
\newblock \emph{CoRR}, abs/2406.06391.

\bibitem[{Zhong et~al.(2023)Zhong, Guo, Gao, and Wang}]{MemoryBank}
Zhong, W.; Guo, L.; Gao, Q.; and Wang, Y. 2023.
\newblock MemoryBank: Enhancing Large Language Models with Long-Term Memory.
\newblock \emph{arXiv preprint arXiv:2305.10250}.

\bibitem[{Zhong, Lei, and Chen(2022)}]{DBLP:conf/emnlp/ZhongLC22}
Zhong, Z.; Lei, T.; and Chen, D. 2022.
\newblock Training Language Models with Memory Augmentation.
\newblock In Goldberg, Y.; Kozareva, Z.; and Zhang, Y., eds., \emph{Proceedings of the 2022 Conference on Empirical Methods in Natural Language Processing, {EMNLP} 2022, Abu Dhabi, United Arab Emirates, December 7-11, 2022}, 5657--5673. Association for Computational Linguistics.

\end{thebibliography}
\end{document}